\documentclass[conference,letterpaper,10pt]{IEEEtran}
\usepackage{cite}
\usepackage{caption}
\usepackage{subcaption} 
\usepackage{graphicx}
\usepackage{amsmath}
\usepackage{algorithm}
\usepackage{algorithmic}
\graphicspath{ {images/} }
\usepackage{multirow}	

\usepackage{wasysym}
\usepackage{pifont}
\usepackage{color}

\renewcommand \thetable{\arabic{table}}

\begin{document}


\title{\huge Cross-Layer Strategic Ensemble Defense \\ Against Adversarial Examples}



\author{\IEEEauthorblockN{\mbox{Wenqi Wei, Ling Liu, Margaret Loper, Ka Ho Chow, Emre Gursoy, Stacey Truex, Yanzhao Wu}}
\IEEEauthorblockA{School of Computer Science\\
Georgia Institute of Technology, 
Atlanta, Georgia, USA, 30332
\\ Email: wenqiwei@gatech.edu, ling.liu@cc.gatech.edu, margaret.loper@gtri.gatech.edu,\\ \{khchow, memregursoy,  staceytruex, yanzhaowu\}@gatech.edu}
}


\maketitle


\begin{abstract}
Deep neural network (DNN) has demonstrated its success in multiple domains. However, DNN models are inherently vulnerable to adversarial examples, which are generated by adding adversarial perturbations to benign inputs to fool the DNN model to misclassify. In this paper, we present a cross-layer strategic ensemble framework and a suite of robust defense algorithms, which are attack-independent, and capable of auto-repairing and auto-verifying the target model being attacked. Our strategic ensemble approach makes three original contributions. First, we employ input-transformation diversity to design the input-layer strategic transformation ensemble algorithms. Second, we utilize model-disagreement diversity to develop the output-layer strategic model ensemble algorithms. Finally, we create an input-output cross-layer strategic ensemble defense that strengthens the defensibility by combining diverse input transformation based model ensembles with diverse output verification model ensembles. Evaluated over 10 attacks on ImageNet dataset, we show that our strategic ensemble defense algorithms can achieve high defense success rates and are more robust with high attack prevention success rates and low benign false negative rates, compared to existing representative defense methods.  
\end{abstract}


\section{Introduction}

Deep learning has increasingly become ubiquitous in Cloud offerings, Internet of Things, and cyber-physical systems. However, deep neural networks (DNNs) are vulnerable to adversarial examples~\cite{goodfellow6572explaining}, which are artifacts generated by adding human-imperceptible distortions to the benign inputs to fool the target DNN model to misclassify randomly or purposefully. A growing number of attacks has been reported in the literature to generate adversarial examples of varying sophistication. As more defense methods are being proposed, the attack-defense arms race has accelerated the development of more aggressive attacks, and developing effective defenses are shown to be substantially harder than designing new attacks~\cite{goodfellow2018defense,papernot2018sok, biggio2018wild}. How to protect deep learning systems against adversarial input attacks has become a pressing challenge.  

In this paper we present a cross-layer strategic ensemble defense approach with three original contributions. First, we develop the input transformation based ensemble algorithms by leveraging diverse input noise reduction techniques. Second, we develop the output model ensemble algorithms by utilizing model-disagreement diversity to create multiple failure independent model verifiers. Third, we create an input-output cross-layer strategic ensemble defense method that strengthens the robustness of our cross-layer ensemble by combining diverse input transformation ensembles with diverse output model ensembles. Due to the space limit, we only evaluate 10 representative attacks on ImageNet dataset. The results show that our cross-layer strategic ensemble defense can achieve high defense success rates, and is more robust with high attack prevention success rates and low benign false negative rates, compared to existing representative defense approaches.

\section{Overview}
\subsection{Characterization of Adversarial Attacks}
Adversarial examples can be generated by the black-box access to the prediction API of the target model being attacked~\cite{goodfellow6572explaining}. 
In this paper, we measure the adversarial effect of attacks by the attack success rate, the misclassification rate, and the attack confidence, and measure the cost of an attack by the perturbation distance, the perception distance, and the average time to generate one adversarial example.
		
\textbf{Attack Success Rate (ASR)} is defined as the percentage of successful adversarial examples over all attack inputs. 

\textbf{Misclassification Rate (MR)} is defined as the percentage of misclassified adversarial examples over all attack inputs. 

\textbf{Mean confidence on adversarial class (AdvConf)} is defined as the mean confidence on the adversarial class of successful adversarial examples number of successful adversarial examples.

\textbf{Perturbation Distance Cost (DistPerturb)} is defined by the root mean square deviation between benign input $x$ and its adversarial counterpart $x_{adv}$.

\textbf{Perception Distance Cost (DistPercept)} measures the perception distance of successful adversarial examples by applying the human perception distance metrics in \cite{luo2018towards}. 

\textbf{Time cost (Time)}: The per-example generation time for each adversarial attack is measured in seconds.

Table~\ref{table:attacks_results} shows the experimental results of 12 attacks on the ImageNet, containing 1.2 million training images and 50000 validation images in 1000 classes. All experiments are conducted on an Intel 4 core i5-7200U CPU@2.50GHz server with a Nvidia Geforce 1080Ti GPU. We select a pre-trained model with a competitive prediction accuracy for the dataset, and target model trained on ImageNet using TensorFlow MobileNet has a validation accuracy of 0.695. 
We consider queries with pre-generated adversarial examples by  building our attack system on top of the EvadeML-Zoo~\cite{xu2017feature}. The first 100 correctly predicted benign examples in the validation set are selected for the attack experiments. We consider a total of 12 representative attacks, two of which are untargeted attacks (UA): Fast Gradient Sign Method (FGSM~\cite{goodfellow6572explaining}), Basic Iterative Method(BIM~\cite{kurakin2016physical}). The other ten are targeted attacks from five attack algorithms: targeted FGSM(TFGSM), targeted BIM (TBIM), and Carlini $\&$ Wagner attacks (CW$_\infty$, CW$_2$, CW$_0$ \cite{carlini2017towards}). 
$L_\infty$, $L_2$, and $L_0$ are the three perturbation norms. We use two representative types of attack targets: the most-likely attack class in the prediction vector ($y^T = \arg \max \nolimits_{y \neq C_x} \overrightarrow y$, most) and the least-likely attack class ($y^T = \arg \min \overrightarrow y$, LL). $C_x$ is the correct class for input $x$. 
We run experiments of all 12 attacks on ImageNet. The $\theta$ is set to 0.0078 for FGSM as small $\theta$ is sufficient to cause high attack SR. For BIM, the per step $\theta$ is 0.002, and the maximum  $\theta$ is 0.004. For CW attacks, the attack confidence is set to 5 and the maximum optimization iteration is set to 1000. The adversarial example is fed into the target ML model every 100 iterations of optimization to check if the attack is successful. 
We make three observations. (1) Attacks that have higher ASR, though may take longer time, do not directly correlate to the amount of perturbation distortion in distance and perception. (2) Given the same attack target model, for CW and JSMA attacks, the least likely (LL) target cost more time and larger distortions, but this is not true for other attacks, showing the divergence behavior of different attack methods~\cite{wei2018adversarial,liu2019deep}. (3) All attacks also exhibit certain divergence behavior in terms of attack effect. Two examples of the same class under the same attack often result in two diverse destination classes (untargeted) or one successful and one failed (targeted). For the same attack method, the divergence of attack effect varies notably for the classifier trained using different DNN models over the same training dataset. 


\begin{table}[ht]
\centering
\scalebox{0.76}{
\small{
\begin{tabular}{cccccccc}
\hline
\multicolumn{2}{|c|}{ImageNet}                                                          & \multicolumn{2}{c|}{attack effect}                     & \multicolumn{1}{c|}{confidence}                           & \multicolumn{3}{c|}{cost}                                                                           \\ \hline
\multicolumn{2}{|c|}{attack}                                                            & \multicolumn{1}{c|}{ASR} & \multicolumn{1}{c|}{MR}   & \multicolumn{1}{c|}{AdvConf} & \multicolumn{1}{c|}{DistPerturb} & \multicolumn{1}{c|}{DistPercept} & \multicolumn{1}{c|}{Time(s)} \\ \hline
\multicolumn{1}{|c|}{FGSM}                   & \multicolumn{1}{c|}{\multirow{2}{*}{UA}} & \multicolumn{1}{c|}{0.99}        & \multicolumn{1}{c|}{0.99}    & \multicolumn{1}{c|}{0.6408}      & \multicolumn{1}{c|}{1.735}       & \multicolumn{1}{c|}{\textbf{1152}}    & \multicolumn{1}{c|}{0.019}    \\ \cline{1-1} \cline{3-8} 
\multicolumn{1}{|c|}{BIM}                    & \multicolumn{1}{c|}{}                    & \multicolumn{1}{c|}{1}       & \multicolumn{1}{c|}{1}   & \multicolumn{1}{c|}{0.9971}      & \multicolumn{1}{c|}{1.186}       & \multicolumn{1}{c|}{502.5}       & \multicolumn{1}{c|}{0.185}    \\ \hline
\multicolumn{1}{|c|}{}                       & \multicolumn{1}{c|}{LL}                  & \multicolumn{1}{c|}{0}         & \multicolumn{1}{c|}{0.91}   & \multicolumn{1}{c|}{N/A}            & \multicolumn{1}{c|}{N/A}            & \multicolumn{1}{c|}{N/A}            & \multicolumn{1}{c|}{0.22}         \\ \hline
\multicolumn{1}{|c|}{\multirow{2}{*}{TBIM}}  & \multicolumn{1}{c|}{most}                  & \multicolumn{1}{c|}{1}       & \multicolumn{1}{c|}{1}   & \multicolumn{1}{c|}{0.9999}      & \multicolumn{1}{c|}{1.175}       & \multicolumn{1}{c|}{485.4}       & \multicolumn{1}{c|}{0.222}    \\ \cline{2-8} 
\multicolumn{1}{|c|}{}                       & \multicolumn{1}{c|}{LL}                  & \multicolumn{1}{c|}{0.53}        & \multicolumn{1}{c|}{0.79}  & \multicolumn{1}{c|}{0.7263}      & \multicolumn{1}{c|}{1.185}       & \multicolumn{1}{c|}{501.2}       & \multicolumn{1}{c|}{0.338}    \\ \hline
\multicolumn{1}{|c|}{\multirow{2}{*}{CW$_\infty$}} & \multicolumn{1}{c|}{most}                  & \multicolumn{1}{c|}{1}       & \multicolumn{1}{c|}{1}  & \multicolumn{1}{c|}{0.9850}      & \multicolumn{1}{c|}{0.957}       & \multicolumn{1}{c|}{217.9}       & \multicolumn{1}{c|}{74.7}     \\ \cline{2-8} 
\multicolumn{1}{|c|}{}                       & \multicolumn{1}{c|}{LL}                  & \multicolumn{1}{c|}{0.95}        & \multicolumn{1}{c|}{0.96}   & \multicolumn{1}{c|}{0.8155}      & \multicolumn{1}{c|}{1.394}       & \multicolumn{1}{c|}{592.8}       & \multicolumn{1}{c|}{\textbf{237.8}}    \\ \hline
\multicolumn{1}{|c|}{\multirow{2}{*}{CW$_2$}} & \multicolumn{1}{c|}{most}                  & \multicolumn{1}{c|}{1}       & \multicolumn{1}{c|}{1}  & \multicolumn{1}{c|}{0.9069}      & \multicolumn{1}{c|}{0.836}       & \multicolumn{1}{c|}{120.5}       & \multicolumn{1}{c|}{13.2}     \\ \cline{2-8} 
\multicolumn{1}{|c|}{}                       & \multicolumn{1}{c|}{LL}                  & \multicolumn{1}{c|}{0.94}        & \multicolumn{1}{c|}{0.94}   & \multicolumn{1}{c|}{0.7765}      & \multicolumn{1}{c|}{1.021}       & \multicolumn{1}{c|}{204.5}       & \multicolumn{1}{c|}{23.1}     \\ \hline
\multicolumn{1}{|c|}{\multirow{2}{*}{CW$_0$}} & \multicolumn{1}{c|}{most}                  & \multicolumn{1}{c|}{1}       & \multicolumn{1}{c|}{1}  & \multicolumn{1}{c|}{0.97}        & \multicolumn{1}{c|}{\textbf{2.189}}       & \multicolumn{1}{c|}{59.5}        & \multicolumn{1}{c|}{\textbf{662.7}}    \\ \cline{2-8} 
\multicolumn{1}{|c|}{}                       & \multicolumn{1}{c|}{LL}                  & \multicolumn{1}{c|}{1}       & \multicolumn{1}{c|}{1}  & \multicolumn{1}{c|}{0.8056}      & \multicolumn{1}{c|}{\textbf{3.007}}       & \multicolumn{1}{c|}{207.5}       & \multicolumn{1}{c|}{\textbf{794.9}}    \\ \hline
\end{tabular}
}}
\caption{\small Evaluation of attacks. }
\label{table:attacks_results}
\vspace{-0.4cm}
\end{table}

\begin{table*}[ht]
\centering
\scalebox{0.80}{
\small{
\begin{tabular}{ccccccccccccc}
\hline
\multicolumn{1}{|c|}{}&
\multicolumn{1}{c|}{\multirow{2}{*}{Attack}} & \multicolumn{1}{c|}{\multirow{2}{*}{test set}} & \multicolumn{1}{c|}{FGSM} & \multicolumn{1}{c|}{BIM} &  \multicolumn{2}{c|}{TBIM} & \multicolumn{2}{c|}{CW$_\infty$} & \multicolumn{2}{c|}{CW$_2$} & \multicolumn{2}{c|}{CW$_0$}  \\ \cline{4-13}
\multicolumn{1}{|c|}{}&\multicolumn{1}{c|}{} & \multicolumn{1}{c|}{} & \multicolumn{2}{c|}{UA} & \multicolumn{1}{c|}{most} & \multicolumn{1}{c|}{LL} & \multicolumn{1}{c|}{most} & \multicolumn{1}{c|}{LL} & \multicolumn{1}{c|}{most} & \multicolumn{1}{c|}{LL} & \multicolumn{1}{c|}{most} & \multicolumn{1}{c|}{LL}  \\ \hline
\multicolumn{1}{|c|}{\multirow{13}{*}{\rotatebox{90}{ImageNet}}}&\multicolumn{1}{c|}{no defense} & \multicolumn{1}{c|}{0.695} & \multicolumn{1}{c|}{0.01} & \multicolumn{1}{c|}{0} &  \multicolumn{1}{c|}{0} & \multicolumn{1}{c|}{0.21} & \multicolumn{1}{c|}{0} & \multicolumn{1}{c|}{0.04} & \multicolumn{1}{c|}{0} & \multicolumn{1}{c|}{0.6} & \multicolumn{1}{c|}{0} & \multicolumn{1}{c|}{0}   \\ \cline{2-13}
\multicolumn{1}{|c|}{}&\multicolumn{1}{c|}{quan-1-bit} & \multicolumn{1}{c|}{0.24} & \multicolumn{1}{c|}{0.28} & \multicolumn{1}{c|}{0.26} & \multicolumn{1}{c|}{0.29} & \multicolumn{1}{c|}{0.3} & \multicolumn{1}{c|}{0.27} & \multicolumn{1}{c|}{0.28} & \multicolumn{1}{c|}{0.27} & \multicolumn{1}{c|}{0.29} & \multicolumn{1}{c|}{0.28} & \multicolumn{1}{c|}{0.22}  \\ \cline{2-13}
\multicolumn{1}{|c|}{}&\multicolumn{1}{c|}{quan-4-bit} & \multicolumn{1}{c|}{0.695} & \multicolumn{1}{c|}{0.05} & \multicolumn{1}{c|}{0.05} & \multicolumn{1}{c|}{0.08} & \multicolumn{1}{c|}{0.83} & \multicolumn{1}{c|}{0.31} & \multicolumn{1}{c|}{0.76} & \multicolumn{1}{c|}{0.47} & \multicolumn{1}{c|}{0.82} & \multicolumn{1}{c|}{0.1} & \multicolumn{1}{c|}{0.58} \\ \cline{2-13}
\multicolumn{1}{|c|}{}&\multicolumn{1}{c|}{medfilter-2*2} & \multicolumn{1}{c|}{0.65} & \multicolumn{1}{c|}{0.22} & \multicolumn{1}{c|}{0.28} & \multicolumn{1}{c|}{0.37} & \multicolumn{1}{c|}{0.78}& \multicolumn{1}{c|}{0.68} & \multicolumn{1}{c|}{0.82} & \multicolumn{1}{c|}{0.74} & \multicolumn{1}{c|}{0.85} & \multicolumn{1}{c|}{0.84} & \multicolumn{1}{c|}{0.85}\\ \cline{2-13}
\multicolumn{1}{|c|}{}&\multicolumn{1}{c|}{medfilter-3*3} & \multicolumn{1}{c|}{0.61} & \multicolumn{1}{c|}{0.33} & \multicolumn{1}{c|}{0.41} & \multicolumn{1}{c|}{0.51} & \multicolumn{1}{c|}{0.73} & \multicolumn{1}{c|}{0.7} & \multicolumn{1}{c|}{0.8} & \multicolumn{1}{c|}{0.73} & \multicolumn{1}{c|}{0.78} & \multicolumn{1}{c|}{0.79} & \multicolumn{1}{c|}{0.82} \\\cline{2-13}
\multicolumn{1}{|c|}{}&\multicolumn{1}{c|}{NLM-11-3-2} & \multicolumn{1}{c|}{0.7} & \multicolumn{1}{c|}{0.05} & \multicolumn{1}{c|}{0.09} & \multicolumn{1}{c|}{0.1} & \multicolumn{1}{c|}{0.82}& \multicolumn{1}{c|}{0.28} & \multicolumn{1}{c|}{0.75} & \multicolumn{1}{c|}{0.43} & \multicolumn{1}{c|}{0.87} & \multicolumn{1}{c|}{0.04} & \multicolumn{1}{c|}{0.25} \\ \cline{2-13}
\multicolumn{1}{|c|}{}&\multicolumn{1}{c|}{NLM-11-3-4} & \multicolumn{1}{c|}{0.66} & \multicolumn{1}{c|}{0.1} & \multicolumn{1}{c|}{0.25} & \multicolumn{1}{c|}{0.3} & \multicolumn{1}{c|}{0.83} & \multicolumn{1}{c|}{0.59} & \multicolumn{1}{c|}{0.83} & \multicolumn{1}{c|}{0.68} & \multicolumn{1}{c|}{0.84} & \multicolumn{1}{c|}{0.2} & \multicolumn{1}{c|}{0.5}\\ \cline{2-13}
\multicolumn{1}{|c|}{}&\multicolumn{1}{c|}{NLM-13-3-2} & \multicolumn{1}{c|}{0.7} & \multicolumn{1}{c|}{0.06} & \multicolumn{1}{c|}{0.09}& \multicolumn{1}{c|}{0.1} & \multicolumn{1}{c|}{0.81} & \multicolumn{1}{c|}{0.32} & \multicolumn{1}{c|}{0.75} & \multicolumn{1}{c|}{0.46} & \multicolumn{1}{c|}{0.87} & \multicolumn{1}{c|}{0.04} & \multicolumn{1}{c|}{0.27} \\ \cline{2-13}
\multicolumn{1}{|c|}{}&\multicolumn{1}{c|}{NLM-13-3-4} & \multicolumn{1}{c|}{0.665} & \multicolumn{1}{c|}{0.11} & \multicolumn{1}{c|}{0.26}  & \multicolumn{1}{c|}{0.31} & \multicolumn{1}{c|}{0.82}  & \multicolumn{1}{c|}{0.59} & \multicolumn{1}{c|}{0.84} & \multicolumn{1}{c|}{0.68} & \multicolumn{1}{c|}{0.87} & \multicolumn{1}{c|}{0.2} & \multicolumn{1}{c|}{0.51}  \\ \cline{2-13}
\multicolumn{1}{|c|}{}&\multicolumn{1}{c|}{rotation\_-12} & \multicolumn{1}{c|}{0.62} & \multicolumn{1}{c|}{0.41} & \multicolumn{1}{c|}{0.55} & \multicolumn{1}{c|}{0.69} & \multicolumn{1}{c|}{0.77}& \multicolumn{1}{c|}{0.74} & \multicolumn{1}{c|}{0.74} & \multicolumn{1}{c|}{0.73} & \multicolumn{1}{c|}{0.78} & \multicolumn{1}{c|}{0.68} & \multicolumn{1}{c|}{0.64}  \\ \cline{2-13}
\multicolumn{1}{|c|}{}&\multicolumn{1}{c|}{rotation\_-9} & \multicolumn{1}{c|}{0.635} & \multicolumn{1}{c|}{0.39} & \multicolumn{1}{c|}{0.53} & \multicolumn{1}{c|}{0.68} & \multicolumn{1}{c|}{0.8} & \multicolumn{1}{c|}{0.77} & \multicolumn{1}{c|}{0.79} & \multicolumn{1}{c|}{0.78} & \multicolumn{1}{c|}{0.8} & \multicolumn{1}{c|}{0.69} & \multicolumn{1}{c|}{0.72} \\ \cline{2-13}
\multicolumn{1}{|c|}{}&\multicolumn{1}{c|}{rotation\_3} & \multicolumn{1}{c|}{0.68} & \multicolumn{1}{c|}{0.29} & \multicolumn{1}{c|}{0.44}& \multicolumn{1}{c|}{0.57} & \multicolumn{1}{c|}{\textbf{0.88}}  & \multicolumn{1}{c|}{0.71} & \multicolumn{1}{c|}{0.83} & \multicolumn{1}{c|}{0.76} & \multicolumn{1}{c|}{0.82} & \multicolumn{1}{c|}{0.65} & \multicolumn{1}{c|}{0.78}  \\ \cline{2-13}
\multicolumn{1}{|c|}{}&\multicolumn{1}{c|}{rotation\_6} & \multicolumn{1}{c|}{0.68} & \multicolumn{1}{c|}{0.33} & \multicolumn{1}{c|}{0.49} & \multicolumn{1}{c|}{0.67} & \multicolumn{1}{c|}{0.82}  & \multicolumn{1}{c|}{0.78} & \multicolumn{1}{c|}{0.85} & \multicolumn{1}{c|}{0.79} & \multicolumn{1}{c|}{0.83} & \multicolumn{1}{c|}{0.72} & \multicolumn{1}{c|}{0.74}  \\ \hline
\multicolumn{2}{|c|}{\begin{tabular}[c]{@{}c@{}}med-3*3, rot\_-9, rot\_6\end{tabular}} & \multicolumn{1}{c|}{\textbf{0.75}} & \multicolumn{1}{c|}{\textbf{0.53}} & \multicolumn{1}{c|}{\textbf{0.64}} &  \multicolumn{1}{c|}{\textbf{0.75}} & \multicolumn{1}{c|}{\textbf{0.96}} & \multicolumn{1}{c|}{\textbf{0.90}} & \multicolumn{1}{c|}{\textbf{0.92}} & \multicolumn{1}{c|}{\textbf{0.89}} & \multicolumn{1}{c|}{\textbf{0.93}} & \multicolumn{1}{c|}{\textbf{0.87}} & \multicolumn{1}{c|}{\textbf{0.89}} \\ \hline
\end{tabular}
}}
\caption{\small Defense accuracy of different input transformation techniques for ImageNet}
\label{table:feature_massaging}
\end{table*}

\subsection{Existing Defenses and Limitations}
Existing defenses are classified into 3 broad categories: adversarial training, gradient masking, and input transformation. 

{\bf Adversarial training} is a class of defense techniques that aim to improve the generalization of a trained model (the target classifier) against known attacks at prediction (test) time by retraining the target model using both benign training set and adversarial examples generated using known attacks~\cite{goodfellow6572explaining}.
The improved robustness is limited to the known adversarial attack algorithms that generate adversarial examples used in training the target classifier~\cite{papernot2018sok}. 

{\bf Gradient masking} refers to the defense techniques that hide gradient information from an adversary, aiming to reduce the sensitivity of a trained model to small changes in input data~\cite{papernot2016distillation}. 

{\bf Input transformation} refers to the defenses that reduce the sensitivity of the target model to small input changes by applying careful noise reduction to the input data before sending it to the target model for prediction by employing some popular image preprocessing techniques like binary filters and median smoothing filters are employed in~\cite{xu2017feature}. 

{\bf Limitations of Existing Defenses.\/} (1) The performance of existing defense methods is sensitive to the magic parameters inherent in their design, such as the percentage of adversarial examples in a batch in adversarial training, the temperature in Defensive Distillation, the detection threshold trained on benign dataset and adversarial inputs in both input transformation and denoising auto-encoder detector defense. Such dataset-specific and/or attack algorithm-specific control parameters make the defense methods non-adaptive \cite{papernot2018sok}. (2) The detection-only methods, though useful to flag the suspicious inputs with high detection success rate, are considered as passive defenses, not able to make the ML component survive its routine function under attack. Such defenses may not be suitable for applications that cannot tolerate real-time interruptions, such as self-driving cars, disease diagnosis.

\subsection{Solution Approach}
We argue that a robust defense should be attack-independent and can generalize over the attack algorithms, and should not depend on finding the attack-specific or dataset-specific control knob (threshold) to distinguish adversarial inputs from benign inputs. We design our strategic ensemble defense algorithms with three objectives. First, our defense algorithms should be attack-independent, capable of auto-repairing and auto-verifying the target model being attacked, and can generalize over different attacks. Second, our defense algorithm should be transparent to the users of the target learning system, with no modification to the application interface (API) used for prediction (testing). Finally, the runtime defense execution at the prediction phase should be efficient to meet the real-time requirements. 
The following metrics are used to measure and compare each defense method. 

\textbf{Prevention Success Rate (PSR)}: The percentage of the adversarial examples that are repaired and correctly classified by the target model under defense. 

\textbf{Detection Success Rate (TSR)}: The percentage of adversarial examples that could not be repaired but are correctly flagged as the attack example by the defense system. 

\textbf{Defense Success Rate (DSR)}: The percentage of adversarial examples that are either repaired or detected. DSR = PSR + TSR.

\textbf{False Positive Rate (FP)}: The percentage of the adversarial examples that can be correctly classified (repaired) but are flagged as adversarial when all inputs are adversarial examples. For the benign test set, we use BFP to represent the percentage of the correctly classified benign examples being flagged as adversarial.

\section{Input Transformation Ensemble}
\label{input}

\subsection{Exploiting Input Denoising Diversity}
\label{denoising}

The main goal of developing input-transformation ensemble methods is to apply data modality specific input noise reduction techniques to clean the input to the target model, aiming to remove adversarial perturbations. We argue that the input noise reduction techniques should be chosen to preserve certain verifiable properties, such as the test accuracy of the target model on benign inputs, while capable of removing the adverse effect from an adversarial input. Image smoothing and image augmentation are common techniques for image noise reduction. The former includes pixel quantization by color bit depth reduction, local spatial smoothing, and non-local spatial smoothing. The latter includes image rotation, image cropping, and rescaling, image quilting and compression. 

\textbf{Rotation:} As a standard image geometric transformation technique provided in SciPy library~\cite{SciPy-Rotation}, rotation preserves the geometric distance of the image and does not change the neighborhood information for most of the pixels except for the corner cases. In our prototype defense, the rotation degree is varied from -12 to 12 with an interval of 3 degree. 

\textbf{Color-depth reduction:} It reduces the color depth of 8 bits ($2^{8}=256$ values) to $i$ bits ($2^{i}$ values). If $i=1$, then the bit quantization will replace the [0,255] space to 1-bit encoding with 2 values: it takes 0 when the nearby pixel value is smaller than 127 and takes 1 when pixel values are in the range of [128, 255]. 
We use quan-$i$-bit to denote the quantization of the input image from the original 8-bit encoded color depth to $i$-bit ($1\le i <8$). 

\textbf{Local spatial smoothing:} It uses nearby pixels to smooth each pixel, with Gaussian, mean or median smoothing~\cite{szeliski2010computer,SciPy-MedianFilter}. A median filter runs a sliding window over each pixel of the image, where the center pixel is replaced by the median value of the neighboring pixels within the window. The size of the window is a configurable parameter, ranging from 1 up to the image size. MedFilter-$k*k$ denotes the median filter with neighborhood kernel size $k*k$. A square shape window size, e.g., $2\times 2$ or $3\times 3$, is often used with reflect padding.

\textbf{Non-local spatial smooth (NLM):} It smooths over similar pixels by exploring a larger neighborhood ($11\times 11$ search window) instead of just nearby pixels and replaces the center patch (say size of $2\times 2$) with the (Gaussian) weighted average of those similar patches in the search window~\cite{buades2005non,OpenCV-Python}. NLM-$a$-$b$-$c$ denotes non-local means smoothing filter with searching window size $a*a$, patch size $b*b$ and Gaussian distribution parameter $c$. NLM 11-2-4 refers to the NLM filter with $11\times 11$ search window, $2\times 2$ patch size and the filter strength of 4.

Unlike adversarial input which injects a small amount of crafted noise to selected pixels in each benign image, input transformation is applied to entire image uniformly, which utilizes the inconsistency of the adversarial examples in terms of the location and the amount of noise to make the perturbation less or no longer effective. 
As the prerequisite for the transformation algorithm selection, all positive examples should be remain positive under different input transformation techniques. Also each negative example tends to be negative in its own way and each adversarial example is destructive in its own way. Hence, different input transformation techniques tend to have different noise reduction effect on adversarial inputs. Strategic ensemble of multiple input transformation techniques can provide robust defense by exploiting noise reduction diversity. 

Assume that we have a pool of diverse candidate transformation techniques of size $m$, by selecting $k$ ($m \geq k\geq 1$) input transformation techniques out of $m$, we obtain $k$ different versions for each input example. The strategic ensemble of $k$ input transformation techniques is to find those that can effectively complement one another on negative examples. A primary criterion for the candidate selection is the high test accuracy on benign test set compared to the test accuracy of the original input example. Time cost could be another criterion. For example, the rotation matrix has a simple and fast transformation. It takes only 0.19s to rotate a color image, compared to 6s for median filter and 59s for non-local filter on the same image. In contrast, generating an adversarial example using CW attack family are order of magnitude more expensive with CW$_2$ at 13$\sim$23 seconds on average to generate an adversarial example and CW$_0$ at 662$\sim$795 seconds per input example on average. 

To verify our analysis, we conduct experiments for the four types of input transformation methods on the benign validation set ImageNet, as well as the adversarial examples generated by all 10 attacks. The benign test accuracy is used as a reference to choose the input transformation method that preserves the competitive test accuracy on the noise reduced version of the benign test set.
Table~\ref{table:feature_massaging} shows the results. We observe several interesting facts. First, employing an input transformation technique can improve the robustness of the target model under attack. However, no single method effective across all 10 attacks. Second, the least-likely attacks are relatively easier to defend than the most-likely attacks. One reason could be that the perturbation in the least-likely attacks is larger and the noise reduction may work more effectively. Finally, strategic ensemble of different input transformation techniques, especially those that have competitive benign test accuracy, can provide good average robustness over the 10 attacks. 

\begin{table*}[ht]
\centering
\scalebox{0.80}{
\small{
\begin{tabular}{|c|c|c|c|c|c|c|c|c|c|c|c|c|c|}
\hline
\multirow{2}{*}{ensemble formation} & \multirow{2}{*}{\begin{tabular}[c]{@{}c@{}}team  strategy\end{tabular}} & FGSM & BIM & \multicolumn{2}{c|}{CW$_\infty$} & \multicolumn{2}{c|}{CW$_2$} & \multicolumn{2}{c|}{CW$_0$} & \multirow{2}{*}{BFP} & \multirow{2}{*}{\begin{tabular}[c]{@{}c@{}} PSR \end{tabular}} & \multirow{2}{*}{\begin{tabular}[c]{@{}c@{}} TSR \end{tabular}} & \multirow{2}{*}{\begin{tabular}[c]{@{}c@{}} DSR \end{tabular}} \\ \cline{3-10}
 &  & \multicolumn{2}{c|}{UA} & most & LL & most & LL & most & LL &  &  &  &  \\ \hline
\begin{tabular}[c]{@{}c@{}}ImageNet (3*3, -9, 6)\end{tabular} & Conf-$L_1$ & \textbf{0.75} & \textbf{0.53}  & \textbf{0.90} & \textbf{0.92} & \textbf{0.89} & \textbf{0.93} & \textbf{0.87} & \textbf{0.89} & \textbf{0.03} & \textbf{0.727} & \textbf{0.108} & \textbf{0.835} \\ \hline
\begin{tabular}[c]{@{}c@{}}ImageNet (3*3, -9, 6)\end{tabular} & $L_1$-upper 1.5 & 0.55 & 0.77  & 0.92 & 0.96 & 0.64 & 0.96 & 0.94 & 1 & 0.1 & 0 & 0.843 & 0.843 \\ \hline
\begin{tabular}[c]{@{}c@{}}ImageNet (3*3, -9, 6)\end{tabular} & $L_1$-lower 0.5 & 0.92 & 0.93  & 1 & 0.99 & 0.9 & 0.99 &  1 & 1 &  0.59 & 0 & 0.966 & 0.966 \\ \hline
\begin{tabular}[c]{@{}c@{}}ImageNet (3*3, -9, 6)\end{tabular} & Adv-Tr 1.693 &  0.479 & 0.633  & 0.778 & 0.979 & 0.727 & 0.976 & 0.891 & 1  & 0.036 & 0 & 0.806 & 0.806 \\ \hline
\begin{tabular}[c]{@{}c@{}}ImageNet (5-bit, 2*2, 11-3-4)\end{tabular} & Adv-Tr 1.254 & 0.5 & 0.429  & 0.756 & 1 & 0.773 & 1 & 0.927 & 1 &  0.055 & 0 & 0.795 & 0.795 \\ \hline
\end{tabular}
}}
\caption{\small Input-layer strategic teaming defense on ImageNet (bold). The detection-only defense using our attack-independent $L_{1}$ thresholds are  included for comparative analysis.}
\label{table:thresholdteaming_new}
\vspace{-0.5cm}
\end{table*}

\begin{figure}[ht]
\centering
\includegraphics[scale=.75]{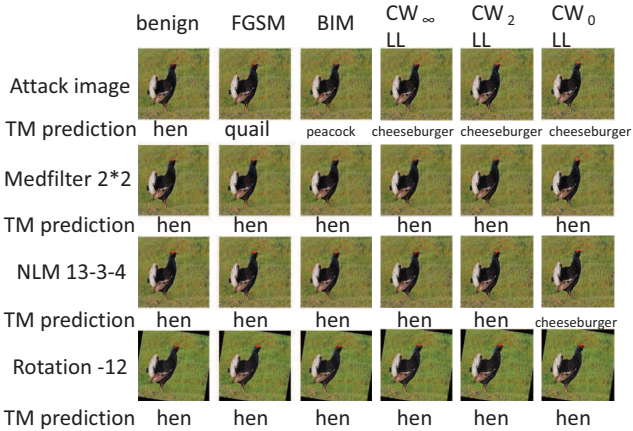}
\vspace{-0.2cm}
\caption{\small ImageNet input-transformation ensemble defense.}
\label{figure:imagenetdefense}
\end{figure}

Table~\ref{table:thresholdteaming_new} shows the input transformation ensemble results on the ImageNet dataset. The first row shows that our strategic teaming algorithm is attack-independent with average DSR of 83.5\% over 10 attacks for ImageNet.  
Next, we compare our results with the design of our {\em attack-independent} version of detection-only defense using the fixed $L_{1}$ upper bound (say $1.5$) or the fixed lower bound (say $0.5$). Conceptually, either an upper bound $L_{1}$ threshold that is above 1 or close to 2, or a lower bound $L_{1}$ threshold that is lower than 1 or close to zero, can be used as the detection threshold.
We compute the $L_{1}$ distance between the prediction vector of each denoised version with the prediction vector of its original input $x$. If the $L_{1}$ distance above the detection-threshold (either upper or lower fixed bound), we flag it as adversarial input. We also include the attack-dependent detection-only method~\cite{xu2017feature}, which uses {\em adversarial example} to set the adversarial detection thresholds. We use the best combo identified in \cite{xu2017feature} in this comparison. We make two observations from Table~\ref{table:thresholdteaming_new}: (1) Adversarial example-based threshold detection on attack examples and benign test set can provide high DSR for the detection-only methods ((DSR=TSR). The adversarial threshold of 1.693, can achieve high DSR of 80.6\% for ImageNet. 
(2) Using attack-independent fixed $L_{1}$ upper or lower bound threshold, adversarial examples can be flagged with reasonable detection success rate (TSR) but at the cost of higher BFP.

Figure~\ref{figure:imagenetdefense} provide an example illustration of three diverse input transformation methods on an ImageNet test example of hen under six scenarios: no attack (benign), two untargeted attacks and three targeted attacks. It shows the effectiveness of the input transformation ensemble for this example input. However, for some examples of ImageNet, these three input transformation techniques may not be as effective as this case, which is one of the primary motivation for us to develop the output verification model ensemble as an alternative joint force to the input transformation ensemble defense.

\section{Output Verification Model Ensemble}
\label{output}

The main objective of the output-layer verification model ensemble is to protect the target model (TM) with the capability to verify and repair the prediction outcome of TM using multiple failure-independent model verifiers and to use the ensemble-approved prediction as the final output of the target model (TM). We exploit the model disagreement diversity to the output-layer strategic model ensembles. {\em First}, we select the baseline candidate models based on a number of criteria, such as high test accuracy, which should be comparable to that of the target model on benign test set, and high model diversity on disagreement measures, such as Kappa($\kappa$)-statistics\cite{mchugh2012interrater} for each pair of the baseline models.
Let $N$ denote the cardinality of the prediction result set, $K$ denote the number of classes, $N_{ij}$ denote the number of instances in the dataset that are labeled as class $i$ by one model and as class $j$ by the other model. $\kappa$ metric is defined as:
\begin{align}
    f_{\kappa} &= \frac{{\frac{{\sum\nolimits_{i = 1}^K {{N_{ii}}} }}{N} - \sum\nolimits_{i = 1}^K {(\frac{{{N_{i*}}}}{N} - \frac{{{N_{*i}}}}{N})} }}{{1 - \sum\nolimits_{i = 1}^K {(\frac{{{N_{i*}}}}{N} - \frac{{{N_{*i}}}}{N})} }} \label{equa:kappa} 
\end{align}
In Equation~\ref{equa:kappa}, $\frac{{\sum\nolimits_{i = 1}^K {{N_{ii}}}}}{N}$ denotes the agreement percentage, i.e., the percentage of agreement made by the two classifiers $i$ and $j$ under the same series of queries. $\sum\nolimits_{i = 1}^K (\frac{{{N_{i*}}}}{N} - \frac{{{N_{*i}}}}{N})$ denotes the chance agreement in which the $*$ is any label in the output space. The $\kappa$ metric is pair-wise metric. The closer the $\kappa$ metric is to 1, the more agreements are made by the two models. The closer the $\kappa$ metric is to 0, the more diverse 
the two models are in terms of disagreement. 

The baseline candidate verification models are selected first based on their test accuracy. For ImageNet, MobileNet is the target model, we select four pre-trained DNN classifiers: VGG-16, VGG-19, ResNet-50 and Inception-V3 as the baseline candidate models for illustration in this paper. We compute the Kappa($\kappa$) for each pair of the candidate models in the baseline model pool of five models: the target model (TM) and four defense models (DMs). Then we build the kappa-ranked list of Kappa-diverse ensembles by the increasing order of the average pairwise Kappa value for each team. For a pool of 5 models, the total combination of ensembles of size 3 or higher is 60 (5$\times$4$\times$3). To avoid low quality ensemble that has low diversity, we select the top $3$ Kappa team by removing those ensembles that have high $\kappa$ value. We get the top three most diverse ensemble teams as the diverse ensemble pool, which are the ensemble teams with the top three lowest average pairwise Kappa values: (DM 1, 3); (DM 1, 3, 4), and (DM 1, 2, 3, 4).

\begin{table*}[ht]
\centering
\scalebox{0.80}{
\small{
\begin{tabular}{ccccccccccccc}
\hline
\multicolumn{1}{|c|}{}&\multicolumn{1}{c|}{\multirow{2}{*}{model}}        & \multicolumn{1}{c|}{\multirow{2}{*}{\begin{tabular}[c]{@{}c@{}}benign \\ acc\end{tabular}}}                                      & \multicolumn{1}{c|}{FGSM} & \multicolumn{1}{c|}{BIM}  &  \multicolumn{2}{c|}{TBIM}                             & \multicolumn{2}{c|}{CW$_\infty$}                      & \multicolumn{2}{c|}{CW$_2$}                           & \multicolumn{2}{c|}{CW$_0$}                                                       \\ \cline{4-13}
\multicolumn{1}{|c|}{}&\multicolumn{1}{c|}{}          & \multicolumn{1}{c|}{}                                                            & \multicolumn{2}{c|}{UA}       & \multicolumn{1}{c|}{most}    & \multicolumn{1}{c|}{LL}  &   \multicolumn{1}{c|}{most}   & \multicolumn{1}{c|}{LL}   & \multicolumn{1}{c|}{most}   & \multicolumn{1}{c|}{LL}   & \multicolumn{1}{c|}{most}   & \multicolumn{1}{c|}{LL}                         \\ \hline
\multicolumn{1}{|c|}{\multirow{11}{*}{\rotatebox{90}{ImageNet}}}&\multicolumn{1}{c|}{TM}     & \multicolumn{1}{c|}{0.695}                                                             & \multicolumn{1}{c|}{0.01} & \multicolumn{1}{c|}{0}    & \multicolumn{1}{c|}{0}     & \multicolumn{1}{c|}{0.21}   & \multicolumn{1}{c|}{0}    & \multicolumn{1}{c|}{0.04} & \multicolumn{1}{c|}{0}    & \multicolumn{1}{c|}{0.06} & \multicolumn{1}{c|}{0}    & \multicolumn{1}{c|}{0}     \\ \cline{2-13}
\multicolumn{1}{|c|}{}&\multicolumn{1}{c|}{DM 1}  & \multicolumn{1}{c|}{0.67}                                                                 & \multicolumn{1}{c|}{0.73} & \multicolumn{1}{c|}{0.77}  & \multicolumn{1}{c|}{0.82}  & \multicolumn{1}{c|}{0.82}   & \multicolumn{1}{c|}{0.81} & \multicolumn{1}{c|}{0.81} & \multicolumn{1}{c|}{0.81} & \multicolumn{1}{c|}{0.82} & \multicolumn{1}{c|}{0.8}  & \multicolumn{1}{c|}{0.79}              \\ \cline{2-13}
\multicolumn{1}{|c|}{}&\multicolumn{1}{c|}{DM 2}   & \multicolumn{1}{c|}{0.68}                                                               & \multicolumn{1}{c|}{0.7}  & \multicolumn{1}{c|}{0.78}  & \multicolumn{1}{c|}{0.81}  & \multicolumn{1}{c|}{0.85}    & \multicolumn{1}{c|}{0.83} & \multicolumn{1}{c|}{0.83} & \multicolumn{1}{c|}{0.84} & \multicolumn{1}{c|}{0.84} & \multicolumn{1}{c|}{0.81} & \multicolumn{1}{c|}{0.76}               \\ \cline{2-13}
\multicolumn{1}{|c|}{}&\multicolumn{1}{c|}{DM 3}    & \multicolumn{1}{c|}{0.67}                                                              & \multicolumn{1}{c|}{0.78} & \multicolumn{1}{c|}{0.84} & \multicolumn{1}{c|}{0.84}  & \multicolumn{1}{c|}{0.84}   & \multicolumn{1}{c|}{0.85} & \multicolumn{1}{c|}{0.83} & \multicolumn{1}{c|}{0.83} & \multicolumn{1}{c|}{0.84} & \multicolumn{1}{c|}{0.83} & \multicolumn{1}{c|}{0.8}               \\ \cline{2-13}
\multicolumn{1}{|c|}{}&\multicolumn{1}{c|}{DM 4}     & \multicolumn{1}{c|}{0.735}                                                             & \multicolumn{1}{c|}{0.86} & \multicolumn{1}{c|}{0.85}  & \multicolumn{1}{c|}{0.91}  & \multicolumn{1}{c|}{0.93} & \multicolumn{1}{c|}{0.92} & \multicolumn{1}{c|}{0.91} & \multicolumn{1}{c|}{0.92} & \multicolumn{1}{c|}{0.9}  & \multicolumn{1}{c|}{0.91} & \multicolumn{1}{c|}{0.84}              \\ \cline{2-13}
\multicolumn{1}{|c|}{}&\multicolumn{1}{c|}{\begin{tabular}[c]{@{}c@{}}RandBase: DM 1,2,3\end{tabular}}     & \multicolumn{1}{c|}{0.770}     & \multicolumn{1}{c|}{0.92} & \multicolumn{1}{c|}{0.92} &  \multicolumn{1}{c|}{0.91}  & \multicolumn{1}{c|}{0.93}   & \multicolumn{1}{c|}{0.92} & \multicolumn{1}{c|}{0.94} & \multicolumn{1}{c|}{0.91} & \multicolumn{1}{c|}{0.93} & \multicolumn{1}{c|}{0.92} & \multicolumn{1}{c|}{0.95}                 \\ \cline{2-13}
\multicolumn{1}{|c|}{}&\multicolumn{1}{c|}{\begin{tabular}[c]{@{}c@{}}Rand$\kappa$: DM 1,2,3,4\end{tabular}}     & \multicolumn{1}{c|}{0.755}  & \multicolumn{1}{c|}{0.83} & \multicolumn{1}{c|}{0.9}  & \multicolumn{1}{c|}{0.92}  & \multicolumn{1}{c|}{0.92}     & \multicolumn{1}{c|}{0.92} & \multicolumn{1}{c|}{0.91} & \multicolumn{1}{c|}{0.92} & \multicolumn{1}{c|}{0.9}  & \multicolumn{1}{c|}{0.89} & \multicolumn{1}{c|}{0.89}                   \\ \cline{2-13}
\multicolumn{1}{|c|}{}&\multicolumn{1}{c|}{\begin{tabular}[c]{@{}c@{}}Best$\kappa$: DM 1,3,4\end{tabular}}       & \multicolumn{1}{c|}{0.805}   & \multicolumn{1}{c|}{\textbf{0.94}} & \multicolumn{1}{c|}{0.93}  & \multicolumn{1}{c|}{0.97}  & \multicolumn{1}{c|}{0.95}  & \multicolumn{1}{c|}{0.96} & \multicolumn{1}{c|}{0.96} & \multicolumn{1}{c|}{\textbf{0.95}} & \multicolumn{1}{c|}{0.95} & \multicolumn{1}{c|}{0.91} & \multicolumn{1}{c|}{0.97}               \\ \cline{2-13}
\multicolumn{1}{|c|}{}&\multicolumn{1}{c|}{rot\_6 $\rightarrow$ RandBase}  & \multicolumn{1}{c|}{0.785} & \multicolumn{1}{c|}{0.93} & \multicolumn{1}{c|}{0.9}  & \multicolumn{1}{c|}{0.92} & \multicolumn{1}{c|}{0.95} & \multicolumn{1}{c|}{0.93}  & \multicolumn{1}{c|}{0.94} & \multicolumn{1}{c|}{0.91} & \multicolumn{1}{c|}{0.95} & \multicolumn{1}{c|}{0.89}  & \multicolumn{1}{c|}{0.94}                \\ \cline{2-13}
\multicolumn{1}{|c|}{}&\multicolumn{1}{c|}{rot\_6 $\rightarrow$ Rand$\kappa$}   & \multicolumn{1}{c|}{0.745}  & \multicolumn{1}{c|}{0.85} & \multicolumn{1}{c|}{0.87}  &  \multicolumn{1}{c|}{0.88} & \multicolumn{1}{c|}{0.87}   & \multicolumn{1}{c|}{0.87} & \multicolumn{1}{c|}{0.85} & \multicolumn{1}{c|}{0.86} & \multicolumn{1}{c|}{0.88} & \multicolumn{1}{c|}{0.89} & \multicolumn{1}{c|}{0.88}  \\ \cline{2-13}
\multicolumn{1}{|c|}{}&\multicolumn{1}{c|}{rot\_6 $\rightarrow$ Best$\kappa$}  & \multicolumn{1}{c|}{0.825} & \multicolumn{1}{c|}{0.89} & \multicolumn{1}{c|}{0.94} &  \multicolumn{1}{c|}{0.96} & \multicolumn{1}{c|}{0.96} &  \multicolumn{1}{c|}{0.96} & \multicolumn{1}{c|}{0.93} & \multicolumn{1}{c|}{0.93} & \multicolumn{1}{c|}{\textbf{0.96}} & \multicolumn{1}{c|}{0.96} & \multicolumn{1}{c|}{\textbf{0.98}}    \\     \cline{2-13}     
\multicolumn{1}{|c|}{}&\multicolumn{1}{c|}{rot\_6 + Best$\kappa$}  & \multicolumn{1}{c|}{0.89} & \multicolumn{1}{c|}{0.89} & \multicolumn{1}{c|}{\textbf{0.96}} &  \multicolumn{1}{c|}{\textbf{1}} & \multicolumn{1}{c|}{\textbf{1}}  & \multicolumn{1}{c|}{\textbf{1}} & \multicolumn{1}{c|}{\textbf{1}} & \multicolumn{1}{c|}{0.93} & \multicolumn{1}{c|}{\textbf{0.96}} & \multicolumn{1}{c|}{\textbf{0.99}} & \multicolumn{1}{c|}{0.97} \\ \hline
\end{tabular}
}}
\caption{\small Prediction Accuracy of the target model(TM) with 10 attacks, the baseline defense model(DM), the random baseline model teaming (RandBase), the random $\kappa$ teaming (Rand$\kappa$) and the Best$\kappa$ teaming for ImageNet}
\label{table:model_massaging}
\end{table*}

To compare the robustness of different model teaming defense algorithms, we conduct a set of experiments using all 10 attacks on ImageNet. We include in our comparison the target model and four individual DMs for ImageNet and three output verification model ensemble teams: random ensemble from the baseline model pool, random $\kappa$ ensemble and the best $\kappa$ ensemble. Table~\ref{table:model_massaging} reports the results. First, we observe that the target model has either zero or close to zero test accuracy. Second, each individual defense model (DM) has higher test accuracy under all 10 attacks untargeted attacks. Third, the test accuracy of DMs under targeted attacks is higher than that under untargeted attack. The reason that the four DM models provide better robustness over all 10 attacks compared to the TM is two folds: (1) The adversarial examples are generated over the black box access to the prediction API of the target model (TM). (2) The adverse effect of these adversarial examples on each of the four defense models (DMs) is due to the transferability of adversarial examples~\cite{papernot2016transferability}. 
Finally, the best$\kappa$ model ensemble is most effective over all 10 attacks in terms of average DSR (test/prediction accuracy), and all three strategic output ensemble teams are more robust against adversarial examples regardless whether it is the random base ensemble, or the random$\kappa$ ensemble from the pool of  $\kappa$ ranked teams, or the Best$\kappa$ ensemble. The last four defense ensembles are formulated by our cross-layer strategic ensemble selection methods, which we discuss in the next section.

\begin{table*}[ht]
\centering
\scalebox{0.80}{
\small{
\begin{tabular}{ccccccccccccc}
\hline
\multicolumn{2}{|c|}{CIFAR-10} & \multicolumn{1}{c|}{FGSM} & \multicolumn{1}{c|}{BIM} &  \multicolumn{2}{c|}{TBIM} &  \multicolumn{2}{c|}{CW$_\infty$} & \multicolumn{2}{c|}{CW$_2$} & \multicolumn{2}{c|}{CW$_0$} \\ \cline{1-12}
\multicolumn{1}{|c|}{Attack} & \multicolumn{1}{c|}{benign acc} & \multicolumn{2}{c|}{UA} & \multicolumn{1}{c|}{most} & \multicolumn{1}{c|}{LL} & \multicolumn{1}{c|}{most} & \multicolumn{1}{c|}{LL} & \multicolumn{1}{c|}{most} & \multicolumn{1}{c|}{LL} & \multicolumn{1}{c|}{most} & \multicolumn{1}{c|}{LL}  \\ \hline
\multicolumn{1}{|c|}{Strategic Ensemble} & \multicolumn{1}{c|}{\textbf{0.9446}} & \multicolumn{1}{c|}{\textbf{0.93}} & \multicolumn{1}{c|}{\textbf{0.99}}  &  \multicolumn{1}{c|}{\textbf{0.94}} & \multicolumn{1}{c|}{\textbf{0.962}}  & \multicolumn{1}{c|}{\textbf{0.97}} & \multicolumn{1}{c|}{\textbf{0.98}} & \multicolumn{1}{c|}{\textbf{0.97}} & \multicolumn{1}{c|}{\textbf{0.99}} & \multicolumn{1}{c|}{\textbf{0.98}} & \multicolumn{1}{c|}{\textbf{1}} \\ \hline
\multicolumn{1}{|c|}{AdvTrain} & \multicolumn{1}{c|}{0.879} & \multicolumn{1}{c|}{0.64} & \multicolumn{1}{c|}{0.58}  &  \multicolumn{1}{c|}{0.464} & \multicolumn{1}{c|}{0.798}  & \multicolumn{1}{c|}{0.68} & \multicolumn{1}{c|}{0.77} & \multicolumn{1}{c|}{0.75} & \multicolumn{1}{c|}{0.79} & \multicolumn{1}{c|}{0.44} & \multicolumn{1}{c|}{0.48}  \\ \hline
\multicolumn{1}{|c|}{DefDistill} & \multicolumn{1}{c|}{0.9118} & \multicolumn{1}{c|}{0.6} & \multicolumn{1}{c|}{0.65}  &  \multicolumn{1}{c|}{0.77} & \multicolumn{1}{c|}{0.904} & \multicolumn{1}{c|}{0.79} & \multicolumn{1}{c|}{0.88} & \multicolumn{1}{c|}{0.86} & \multicolumn{1}{c|}{0.9} & \multicolumn{1}{c|}{0.6} & \multicolumn{1}{c|}{0.69}  \\ \hline
\multicolumn{1}{|c|}{EnsTrans} & \multicolumn{1}{c|}{0.8014} & \multicolumn{1}{c|}{0.23} & \multicolumn{1}{c|}{0.4}   &  \multicolumn{1}{c|}{0.406} & \multicolumn{1}{c|}{0.668} & \multicolumn{1}{c|}{0.56} & \multicolumn{1}{c|}{0.61} & \multicolumn{1}{c|}{0.57} & \multicolumn{1}{c|}{0.61} & \multicolumn{1}{c|}{0.19} & \multicolumn{1}{c|}{0.34} \\ \hline
\multicolumn{1}{l}{} & \multicolumn{1}{l}{} & \multicolumn{1}{l}{} & \multicolumn{1}{l}{} & \multicolumn{1}{l}{} & \multicolumn{1}{l}{} & \multicolumn{1}{l}{} & \multicolumn{1}{l}{} & \multicolumn{1}{l}{} & \multicolumn{1}{l}{}  & \multicolumn{1}{l}{}  & \multicolumn{1}{l}{}\\ \hline
\multicolumn{2}{|c|}{ImageNet}  & \multicolumn{1}{c|}{FGSM} & \multicolumn{1}{c|}{BIM} & \multicolumn{2}{c|}{TBIM} & \multicolumn{2}{c|}{CW$_\infty$} & \multicolumn{2}{c|}{CW$_2$} & \multicolumn{2}{c|}{CW$_0$} \\ \cline{1-12}
\multicolumn{1}{|c|}{Attack} & \multicolumn{1}{c|}{benign acc} & \multicolumn{2}{c|}{UA} & \multicolumn{1}{c|}{most} & \multicolumn{1}{c|}{LL}&  \multicolumn{1}{c|}{most} & \multicolumn{1}{c|}{LL} & \multicolumn{1}{c|}{most} & \multicolumn{1}{c|}{LL} & \multicolumn{1}{c|}{most} & \multicolumn{1}{c|}{LL} \\ \hline
\multicolumn{1}{|c|}{Input Transformation Ensemble} & \multicolumn{1}{c|}{0.75} & \multicolumn{1}{c|}{0.53} & \multicolumn{1}{c|}{0.64} & \multicolumn{1}{c|}{0.75} & \multicolumn{1}{c|}{\textbf{0.96}} & \multicolumn{1}{c|}{0.90} & \multicolumn{1}{c|}{0.92} & \multicolumn{1}{c|}{0.89} & \multicolumn{1}{c|}{0.93} & \multicolumn{1}{c|}{0.87} & \multicolumn{1}{c|}{0.89}  \\ \hline
\multicolumn{1}{|c|}{Output Model Ensemble} & \multicolumn{1}{c|}{0.805} & \multicolumn{1}{c|}{\textbf{0.94}} & \multicolumn{1}{c|}{0.93}  & \multicolumn{1}{c|}{\textbf{0.97}} & \multicolumn{1}{c|}{0.95} &  \multicolumn{1}{c|}{\textbf{0.96}} & \multicolumn{1}{c|}{\textbf{0.96}} & \multicolumn{1}{c|}{\textbf{0.95}} & \multicolumn{1}{c|}{0.95} & \multicolumn{1}{c|}{0.91} & \multicolumn{1}{c|}{0.97}  \\ \hline
\multicolumn{1}{|c|}{Cross-layer Strategic Ensemble} & \multicolumn{1}{c|}{0.825} & \multicolumn{1}{c|}{0.89} & \multicolumn{1}{c|}{\textbf{0.94}} & \multicolumn{1}{c|}{0.96} & \multicolumn{1}{c|}{\textbf{0.96}} &  \multicolumn{1}{c|}{\textbf{0.96}} & \multicolumn{1}{c|}{0.93} & \multicolumn{1}{c|}{0.93} & \multicolumn{1}{c|}{\textbf{0.96}} & \multicolumn{1}{c|}{\textbf{0.96}} & \multicolumn{1}{c|}{\textbf{0.98}}  \\ \hline
\multicolumn{1}{|c|}{EnsemInputTrans} & \multicolumn{1}{c|}{0.715} & \multicolumn{1}{c|}{0.41} & \multicolumn{1}{c|}{0.6} &  \multicolumn{1}{c|}{0.74} & \multicolumn{1}{c|}{0.9} &  \multicolumn{1}{c|}{0.8} & \multicolumn{1}{c|}{0.91} & \multicolumn{1}{c|}{0.82} & \multicolumn{1}{c|}{0.92} & \multicolumn{1}{c|}{0.76} & \multicolumn{1}{c|}{0.86}   \\ \hline
\end{tabular}
}}
\caption{\small Defense success rate comparison of cross-layer strategic ensemble (StrategicEnsemble) with adversarial training(AdvTrain), defensive distillation(DefDistill) and input transformation ensemble(EnsemTrans).}
\label{table:defense_combo_compare_other}
\vspace{-0.4cm}
\end{table*}

\section{Input-Output Cross-Layer Strategic Teaming}
\label{cross-layer}

Our cross-layer strategic ensemble defense method is designed to combine the input-layer transformation ensemble with the output-layer model verification ensemble by maximizing the disagreement diversity (failure independence). 

We use the notation {\em inpu-transformation} $\rightarrow$ {\em output ensemble} to denote the cross-layer strategic ensemble that performs input transformation followed by model ensemble verification. We use the notation {\em input-transformation} $+$ {\em output model ensemble} to denote the second type of cross-layer ensemble co-defense strategy. For example, med 2*2 $+$ Rand$\kappa$ denotes the transformed input of $x$ by med 2*2 filer is sent to only the TM and the original input $x$ is sent to the output-layer model ensemble verification team, which output the cross-layer defense-approved prediction result.  
 
We compare four cross-layer strategic ensemble defense algorithms over all 10 attacks on ImageNet and report the results in the last four rows of Table~\ref{table:model_massaging}. For ImageNet, most of the cross-layer ensemble defense teams performs well compared to the top performing input-layer transformation ensembles and the top performing output-layer model verification ensembles, though in some cases, the output model ensemble along can be more effective, such as RandBase: DM 1,2,3 and best$\kappa$ DM: 1,3,4 under FGSM attack and CW$_2$ attack. This also indicates that the input transformation method rot\_6 may not complement well with the output-layer model ensemble. One of our ongoing research is to investigate good criteria for most robust cross-layer ensemble formation. 

\section{Comparison with Existing Defense Approaches}

We conduct the experiments to compare our strategic ensemble approach with the representative defense methods in three broad categories: Adversarial Training(AdvTrain)~\cite{goodfellow6572explaining}, Defensive Distillation(DefDistill)~\cite{papernot2016distillation} and Input Transformation Ensemble(EnsTrans)~\cite{xie2017mitigating}. 
We did not find pre-trained models on ImageNet with the adversarial training or defensive distillation defense and the Intel 4 core i5-7200U CPU@2.50GHz server with the Nvidia Geforce 1090Ti GPU with 3000+ units were not able to complete the adversarial training or the defensive distillation powered training on ImageNet. Thus, we include CIFAR-10 in this set of comparison experiments. For ImageNet, we only compare the strategic teaming defense with Input Transformation Ensemble.  
Table~\ref{table:defense_combo_compare_other} shows the results. For adversarial training, we use adversarial examples generated from FGSM with random $\theta$ from $[0, 0.0156]$ for CIFAR-10. For defensive distillation, the temperature is set to 50 for CIFAR-10. The input transformation ensemble has two parameters for both datasets: the ensemble size $n$ and the crop size $c$, and it computes multiple randomly cropped-and-padded input image ($n$ times) using the given crop size. In our experiments, the ensemble size is set to 10 for both ImageNet and CIFAR-10, and the crop size is set to 28 for CIFAR-10 and 196 for ImageNet according to the recommended settings in~\cite{xie2017mitigating}. 
This set of experiments shows that (1) our cross-layer strategic ensemble approach consistently outperforms the existing defense approaches over all 10 attacks on both datasets; and (2) for CIFAR-10, our strategic teaming achieves 96.5\% average DSR compared to 73.3\% average DSR by the defensive distillation, the best among the three representative existing defense methods. For ImageNet, our output-layer strategic ensemble achieves 95.1\% average DSR, and our cross-layer strategic ensemble achieves 94.1\%, both are much better than the ensemble input transformation approach (72.8\% average DSR). Even our input-transformation ensemble alone achieves 79.7\% average DSR, compared to 72.8\% average DSR by the ensemble input transformation. This further demonstrates the robustness of our diversity-enhanced strategic ensemble algorithms for defense against adversarial examples.

\section{Conclusion}

We have presented a cross-layer strategic ensemble defense approach by combining input transformation ensembles with output verification model ensembles by promoting and guaranteeing ensemble diversity. Our strategic ensemble approach is attack-independent, generalize well over attack algorithms, and is capable of auto-repairing and auto-verifying the target model being attacked. Evaluated using ImageNet and CIFAR-10 over 10 representative attacks, we show that our cross-layer ensemble defense algorithms can achieve high defense success rates, i.e., high test accuracy in the presence of adversarial attacks, and are more robust compared to existing representative defense methods, with high attack prevention success rates (PSR) and low benign false negative rates (BFP). Our ongoing research continues along two dimensions:(1)Developing theoretical foundation for cross-layer strategic ensemble formulation algorithms with verifiable robustness; and (2) Incorporating new generations of attack algorithms, and new generations of defense methods in the empirical comparison framework. 
 
\subsection*{Acknowledgement} This work is partially sponsored by NSF CISE SaTC grant 1564097 and an IBM faculty award.


\bibliographystyle{IEEEtran}
\bibliography{bibliography}

\end{document}